\documentclass[runningheads]{llncs}
\usepackage{graphicx}
\usepackage{tikz}
\usetikzlibrary{matrix}
\usepackage{wrapfig}
\usepackage{latexsym}
\usepackage{amssymb}
\begin{document}
\title{Revisiting Indirect Ontology Alignment : New Challenging Issues in Cross-Lingual Context}
\titlerunning{\textsc{Cimona}}
\author{Marouen Kachroudi\inst{1}\orcidID{0000-0002-7536-0428}
%\and
%Second Author\inst{2,3}\orcidID{1111-2222-3333-4444} \and
%Third Author\inst{3}\orcidID{2222--3333-4444-5555}
}
\authorrunning{M. Kachroudi}
\institute{Universit\'{e} de Tunis El Manar, Facult\'{e} des Sciences de Tunis\\ Informatique Programmation Algorithmique et Heuristique\\LIPAH-LR 1 ES14, 2092 Tunis, Tunisie\\
\email{marouen.kachroudi@fst.rnu.tn}}
\maketitle              % typeset the header of the contribution
\begin{abstract}
Ontology alignment process is overwhelmingly cited in Knowledge Engineering as a key mechanism aimed at bypassing heterogeneity and reconciling various data sources, represented by ontologies, \textit{i.e.}, the the Semantic Web cornerstone. In such infrastructures and environments, it is inconceivable to assume that all ontologies covering a particular domain of knowledge are aligned in pairs. Moreover, the high performance of alignment approaches is closely related to two factors, \textit{i.e.}, time consumption and machine resource limitations. Thus, good quality alignments are valuable and it would be appropriate to exploit them. Based on this observation, this article introduces a new method of indirect alignment of ontologies in a cross-lingual context. Indeed, the proposed method deals with alignments of multilingual ontologies and implements an
indirect ontology alignment strategy based on a composition and reuse of effective direct alignments. The trigger of the proposed method process is based on alignment algebra which governs the semantics composition of relationships and confidence values. The obtained results, after a thorough and detailed experiment are very encouraging and highlight many positive aspects about the new proposed method.

\end{abstract}

\section{Introduction and Motivations}

Semantic Web active actors publish and share their data sources in their own
respective languages or formalisms. Moreover, the explicitation
of the associated concepts related to a particular domain of interest takes advantage of ontologies, considered as the semantic Web cornerstone \cite{BernersLee11}.
This process has been accelerated due to a few initiatives which encourage all the active participants to make their data available to the public. These actors often publish their data sources in their own respective grasp, in order to make this information inter-operable and accessible to members of other communities \cite{BernersLee11}. In addition, the open and dynamic resources of the semantic Web endow it with a heterogeneous aspect, which reflects at once the formats or
the conceptual varieties of its description. Indeed, the informative volume reachable via the semantic Web stresses needs of techniques guaranteeing the share, reuse and interaction of all resources \cite{suchanek}. Based on the fact that
ontologies are likely to be authored by different actors  using  different  terminologies, structures and natural languages, ontology alignment has emerged as a
way to achieve semantic interoperability \cite{euzenat2013}.
The operation of finding correspondences is called \emph{ontology alignment} and its result is a set of correspondences called an alignment \cite{euzenat2013}.
Similarly to ontologies, alignments have their own life cycle \cite{euzenatsharfe2008}.
Firstly, they are created through an alignment process. Then, they can go through
an iterative procedure for evaluation and probably enhancement. Obviously, these
alignments are subject of extensive evaluation, improvement and finally integrated
across multiple environments before being used by several applications: ontology evolution, producing a set of transitive axioms in order to identify corresponding concepts, interpreting conveyed messages between agents, interpreting data circulating across heterogeneous web services, mediating in query/answer scenarios in peer-to-peer systems and distributed systems or databases. Moreover, the alignment life cycle is closely related to the ontology life cycle: as soon as ontologies evolve, new alignments have to be evolved according to the
ontology evolution. This can be achieved by storing the variations undergone by ontologies
and transforming those variations into an alignment. This can be used to compute new
alignments that will update the previous ones. So, application fields are more and more numerous
and they put in front very specific difficulties.
Moreover, existing alignments investigation allows the
reasoning on the context intersections of various
ontological representations. In this context, the
task of reasoning about overlapping context domains
led to support alignment processing and its content management.
Recent researches in the ontology alignment field has largely
focused on dealing with direct ontology
alignment techniques \cite{HertlingP19,AML2015,AnnaneBAJ16,LYAM++2015}. Worthy to cite, a few researches have
focused in an abbreviated way on indirect alignment scenarios.
However, current direct alignment techniques often rely on
semantic correspondence detection between
ontologies component treated as input.

This paper meets challenges strictly bound at the interoperability level. Indeed, it proposes a new idea about a minimal framework for indirect ontology alignment called \textsc{Cimona}
(Cross-lingual Indirect Matching ON generated Alignments). The main idea is to capitalize on existing alignments expressed in the RDF format and which already exist within such a semantic environment. The main thrust of \textsc{Cimona} stands on the fact that it only deals with alignment files to produce new ones. Furthermore, the new proposed approach uses an external resource to maintain the semantic aspect during existing alignments processing.

In this case, previously existing alignments can be replaced by their composition with the ontology update alignment. As demonstrated by this evolution example case, alignment management can rely on alignments composition, more, there a pressant need to reason about alignments. The introduced effective support must be boosted with substantial annotations allowing contributors and systems to select the adequate alignments based on various criteria.
It should also support permanent storage and identification of alignments in order to reliably use the
existing ones. Analogously, in databases, several systems have been designed for offering a
variety of alignment methods and a library of correspondences \cite{Rahm}. Further, an alignment server \cite{euzenat2005alignmentserver} has been designed with
the aim of providing access to existing alignments, and thus facilitating their reuse.
We emphasize that in the context of collaborative alignment systems the above mentioned needs
are vital. Doing so, we can discern two levels in alignment management: (\emph{i}) the hardware infrastructure
and (\emph{ii}) the framework supporting operations over alignments.
The framework environments may allow alignment edition \cite{Noy2003},
alignment processing \cite{euzenat2005alignmentserver}, alignment sharing \cite{euzenat2005alignmentserver}.
These two levels can be mixed in a single system or exploited separately \cite{euzenat2005alignmentserver}.
One of the challenges in this register is to provide an efficient alignment support infrastructure at the
semantic Web scale, such that tools and, more importantly, applications can rely on it to
share, \emph{i.e.}, publish and  mostly reuse, alignments, and more generally to guarantee interoperability.\\

The remainder of this paper is as follows. Section \ref{key} reviews the existing methods in the field of indirect ontology alignment and defines some terminologies and notations for the rest of this paper. Section \ref{cimona} thoroughly describes the \textsc{Cimona} method, its foundation and its various steps as the main contribution of this work.
Section \ref{etude} reports the experimental encouraging results obtained
with the considered test base. Finally, Section \ref{conclusion} draws the conclusion and sketches future issues of this paper.

\section{Basics and Background} \label{key}

Ontology alignment is considered as an evaluation of the
degrees of resemblance or the differences detected on the considered ontological
entities \cite{Ehrig2007}. Besides, the process of alignment can be defined as
follows: being given two ontologies $\mathcal {O}_1$ and $\mathcal {O}_2$, an alignment between
$\mathcal {O}_1$ and $\mathcal {O}_2$ is a set of correspondences,
(\textit{i.e.}, a quadruplet): $<e_{1}, e_{2}, r, Conf(n)>$, with $e_{1} \in \mathcal {O}_1$
and $e_{2} \in \mathcal {O}_2$, $r$ is a relation between two given entities $e_{1}$ and $e_{2}$,
while $Conf(n)$ represents the confidence level in this relation \cite{euzenat2013}.\\

\subsection{Alignment reuse and management} \label{compcomp}

\medskip

Alignments and ontologies can be referenced by their URI,
which makes possible to reuse alignments and ontologies
already published independently of the definition of a given module.
Moreover, the use of a semantics separating the interpretation
of the ontologies and that of the alignments ensures a good
decoupling of the different languages involved. Thus, it makes
it possible to design verification procedures for mediation
and allows an exchange and a reuse of alignments among
different applications.\\

Alignments, like ontologies, must be accompanied during their
life-cycle phases by the appropriate tools and standards.
These function prerequisites can be implemented
as services, the most relevant of which are:

\begin{itemize}
  \item Matching two ontologies by selecting the algorithm to be used and its parameters (including eventually initial alignment);
  \item Storing an alignment in a persistent location;
  \item Extracting an alignment based on its identifier;
  \item Retrieving alignment metadata such as its identifier that can be used to select a precise alignment;
  \item Finding (stored) alignments between two specific ontologies;
  \item Changing an alignment by adding or removing matches;
  \item Cutting alignments according to a well established threshold;
  \item Generating the code by implementing the transformations of the ontology, translations of data
or axioms bridges from a particular alignment;
\end{itemize}

This functional support must be supplemented with rich metadata enabling users
and systems to select appropriate alignments according to various criteria. It must also
support the permanent storage and identification of alignments so that they can be used in a
reliable way. In the field of databases, several systems have been designed to offer
various alignment methods as well as alignment libraries \cite{Dhamankar2004}.
When an ontology evolves into a new version, it is necessary to update
the instances of this ontology and the alignment it has with the other ontologies.
To this extent, a new alignment between the two versions can be established and used to
generate the transformation of the necessary instance and
link the alignments to the handled ontologies.\\

\begin{wrapfigure}{r}{0.5\textwidth}
\label{glance}

  \begin{center}
  \includegraphics[scale=0.3]{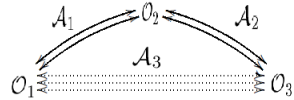}\\
  \caption{Indirect alignment at a glance.}
  \end{center}

\end{wrapfigure}

In a wider context, produced alignments in the semantic Web
can be reused. Indeed, the main advantage of the reuse process reduces the cost of conventional alignment operations. Thus, the basic idea is to deduce new alignments from other existing ones without resorting to deploy needed routine logistics, \emph{i.e.}, treatment and main memory handling. From this
finding, emerged the idea of indirect alignment \cite{euzenatindir}.
In other words, given three ontologies $\mathcal{O}_1$, $\mathcal{O}_2$
and $\mathcal{O}_3$, with $\mathcal{A}_1$ the alignment of $\mathcal{O}_1$
to $\mathcal{O}_2$, $\mathcal{A}_2$ the alignment of $\mathcal{O}_2$ to $\mathcal{O}_3$, then it would be possible to derive a composed alignment $\mathcal{A}_3$ between $\mathcal{O}_1$ and $\mathcal{O}_3$ as sketched by Figure 1. Alignment composition process is mainly based on the alignment algebra proposed by Euzenat \cite{euzenatindir}.\\

In what follows, $e_1$, $e_2$ and $e_3$
are three entities belonging respectively to $\mathcal{O}_1$, $\mathcal{O}_2$
and $\mathcal{O}_3$. Indeed, alignment composition
determines the semantic relations which labels the link between $e_1$ to $e_3$, based on already
existing links between $e_1$, $e_2$ and $e_2$, $e_3$.
Secondly, the composition process determines the confidence value
on the new composed relation between $e_1$ and $e_3$. In fact, relation
composition for two given entities is carried out according to several rules \cite{euzenatindir} recapitulated in Table \ref{algebre},
where ($=$) symbolizes \emph{equivalence}, ($>$) represent \emph{inclusion}, ($<$) stands for \emph{reverse inclusion}, ($\between$) means \emph{overlapping},
and ($\perp$) notify \emph{disjunction}.

\begin{table}
\centering
\caption{Semantic relations composition according to the alignment algebra.}\label{algebre}
\begin{tabular}{|p{1.5cm}|p{1.5cm}|p{1.5cm}|p{1.5cm}|p{1.5cm}|p{1.5cm}|}
\hline
     &   $=$  &  $>$     & $<$           & $\between$   & $\perp$ \\ \hline
$=$  &   $=$  & $>$      & $<$           & $\between$   &   $\perp$   \\ \hline
$>$  &   $>$  & $>$      & $><=\between$ & $>\between$  &   $>\between\perp$     \\ \hline
$<$  &   $<$  & $\Gamma$ & $<$           & $<\between\perp$ &   $\perp$     \\ \hline
$\between$ &    $\between$ & $>\between\perp$  &  $<\between$     & $\Gamma$         &   $>\between\perp$     \\ \hline
$\perp$    &    $\perp$    & $\perp$ & $<\between\perp$ & $<\between\perp$ &   $\Gamma$\\

\hline
\end{tabular}
\end{table}

The composed relation should be supported in turn by a confidence
value ($n$ or $n'$) that reflects the correspondence degree between the two considered
entities. The latter value can be derived according to several ways \cite{euzenatindir} :

%\vspace{-0.35cm}

\begin{itemize}
  \item Multiplication : $Conf(n,n')=n\times n'$
  \item Normalization :   $Conf(n,n')=\frac{n\times n'}{2}$
  \item Maximization : $Conf(n,n')=max(n,n')$
  \item Minimization : $Conf(n,n')=min(n,n')$
\end{itemize}

\subsection{Related work}

\medskip

In the literature, several approaches tried to exploit
and implement the notions previously introduced. These approaches
proposed mechanisms that enable them to connect data fragments
initially isolated. The interconnection operation involves
treatments based on reasoning and/or semantic transitivity into
the considered data amount space.\\

A primitive framework was presented \cite{Embley2004} for
automatically discovering both direct and indirect alignments
over sets of source and target schema elements. In this framework,
multiple techniques are used in a combined way to produce a
final set of alignments. The deployed techniques include terminological
relationship investigation, data-value characteristics, as well as
structural characteristics. It should be emphasized that recognizing
expected data values associated to schema elements and applying
schema structure heuristics are the key ideas to compute indirect alignments.
Indeed, indirect element constituents are detected thanks to five phases, namely \emph{selection},
\emph{union}, \emph{composition}, and \emph{decomposition}
as well as \emph{conversions of named schema element} \cite{Embley2004}.
The latter operations and conversions are mainly and thoroughly
guided by expected confidence values and
structural supervision.\\

Mappings composition has mainly been studied for schemas \cite{Dragut2004,Fagin2005} and
model management \cite{Bernstein2007}. Only a few approaches consider mapping composition
for deriving new mappings in ontology matching. For instance, \cite{Zhang2005} utilizes FMA
as an intermediate ontology to indirectly generate a mapping between FMA and NCI Thesaurus.
\cite{Tordai10} presents an empirical
analysis of mapping composition available in
BioPortal\footnote{https://bioportal.bioontology.org/}.
Authors already studied mapping composition \cite{gross2011} in a previous contribution. The primary focus of this
work was on matching quality (F-measure) by a manual intermediate selection but
not on automatic strategies to select the best intermediates according to their
expected contribution to the overall match quality. Mappings between related ontologies are useful in many ways, in particular
for data integration and enhanced analysis \cite{Noy2009}.\\

Zimmermann's work \cite{zimmermann} has addressed
concisely and formally the alignments composition topic. Alignments
composition heavily rely on a relation algebra based operator. It was designed and
defined in a consistent manner and focusing on the semantics of a given ontology set.
In this context, when handling description logics aligned by simple correspondence,
a procedure for consistency checking over the set of ontologies was established.
This procedure is independent, complete, and operates thanks to local existing
reasoning systems access.\\

A theoretical idea was also presented for building indirect
alignments between multilingual ontologies \cite{Jung09}.
The basic principle of this method is the reuse of already
existing and stored alignment files. An intermediary alignment
should be carried out between source and target ontology to compose
a new alignment using such objects. Beforehand, equivalence
between multilingual entities belonging in both considered distinct
ontologies should be discovered and established by a human expert.
Then, a process of alignment composition is applied using alignment
algebra introduced in \cite{euzenatindir}.\\

Worthy of cite, Gross et \emph{al.} proposed a composition-based
approach for indirectly aligning life science ontologies via one or several
intermediate ontologies \cite{gross2011}. The goal of the proposed solution
is to reuse previously determined ontology alignments towards improved alignment efficiency and quality.
The approach is based on ontology and mapping operators called respectively
\emph{compose}, \emph{match}, \emph{extract} and \emph{merge}. It allows the flexible combination
of several composed alignments and the incremental
extension of alignments by additional cells for
unaligned ontology concepts in some evolutive cases.
In the same register, \cite{Hartung12} investigated a simple approach to generate
mappings between ontologies by reusing and composing existing mappings
across intermediate ontologies. This approach is qualified to be efficient
with highly interconnected ontologies. Consequently, ontologies could be used for composition
and the problem arises to find the appropriate ones providing the
best results. Authors proposed and defined also several measures and strategies to select the
most promising intermediate ontologies for composition.
Also in this register, a framework was proposed \cite{KachroudiISMIS15} to interlink indirectly
ontologies describing the conference organisation domain.\\

To sum up, there are several methods that have addressed the
indirect alignment issue. Each of these methods has its own vision and its own
account for indirect alignment. Furthermore, the driving idea is the reuse of alignments or alignment fragments for many purposes. Indeed, the majority
of approaches have focused on the reuse of existing alignments as a background
knowledge to boost other phases, either alignment or ontology alignment evolution.
In addition, the only work that has methodically addressed this issue is that of Jung
et \emph{al.} \cite{Jung09} although the work remains limited, primitive and without
thorough tests on some concrete cases. In the next section, we introduce \textsc{Cimona}, as the new indirect ontology alignment method.

\section{The proposed method} \label{cimona}

\begin{figure}[htpb] \centering
\includegraphics[width=12.5cm,height=2.5cm]{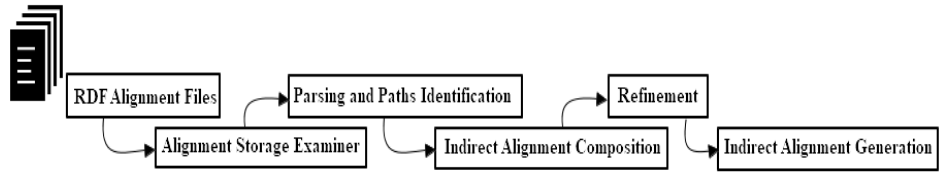}
\caption{The \textsc{Cimona} workflow.}
\label{CCimona}
\end{figure}

In the following, a detailed description of the \textsc{Cimona} method that
revolves around five major components as depicted by Figure \ref{CCimona}.

The global algorithm takes as input a search space consisting
of a finite number of ontology alignment ($\mathcal{A}_1$, $\mathcal{A}_2$,..., $\mathcal{A}_n$). Then, it extracts two direct alignments which will be considered as candidates to alignment composition.\\

\subsection{Alignment Storage Examiner} \label{storageexaminer}

The driving idea is to examine the provided space formed by $n$ ontologies
aligned in pairs, what we call \emph{ontology network}. The purpose of this
module is to determine the number of deducible indirect alignments
from the provided direct ones. The notion of ontology network likens this representation
to a graph, whose nodes are ontologies and arcs are pairwise ontology alignments.
Indeed, this component seeks to check links existence between ($\mathcal {O}_1$, $\mathcal {O}_2$) on a one hand
and between ($\mathcal {O}_2$, $\mathcal {O}_3$) on the other hand.
Thereafter, the rest of the process can be triggered.
These structured alignments conform to two ontologies describing their
domain of knowledge. These ontologies are not necessarily contained in
the reasoning systems because it may suffice to refer to them by their
identifier on the Web, in which case it is possible that several knowledge
bases point to the same ontology. If several of these ontologies are made
accessible at a given time, they can be matched by a specialized system,
for example a server ontology alignment. The alignment is then made available
on the Web. But then, the alignment does not express a priori a
knowledge relative to one or the other of the systems. We can conclude that
an alignment in this context does not express
knowledge from the point of view of a particular ontology or knowledge base.
On the contrary, alignment only affirms relations between two ontologies
without privileging one, since we can not know which ones will be used
by a local knowledge base. In the absence of additional information, it
can then be assumed that an alignment is not directional.
On the other hand, always within this framework, an alignment does not
a priori constitute a matching of local ontologies towards a hypothetical
"global" ontology. An alignment merely defines the relationships between
local terminology, not a terminology that is specific to it.
Thus, alignments are only knowledge referring to ontologies,
but specify inter-ontological knowledge.\\

\subsection{Parsing and Paths Identification} \label{APP}

We call a path, a triple formed by three entities, of which
two equivalent entities and a bridge entity that we define in
the following, in subsection \ref{bridge}. After defining the available direct alignments between ontologies involved in the previous component, the current module initiates the next phase, for parsing and preprocessing. This module operates on the supplied alignment \texttt{rdf} files. This parsing step is performed via Jena\footnote{https://jena.apache.org/documentation/ontology/}
API and returns two ontological entities lists that have been initially matched in pairs. Each list refers to a direct alignment output.
The storage of each list is achieved through entity couples to
which we add the semantic link between them and the value of their correspondence degree. The syntactic preprocessing step relies on the implementation of standard processing techniques to remove empty words, non-alphanumeric characters, lowercasing, etc.

\subsection{Indirect Alignment Composition} \label{AComp}

This module is the kernel of the \textsc{Cimona} method, it
encompasses three complementary sub modules.
Through this pivot module,
it would be possible to produce new composed
indirect alignments from other direct
alignments according to several phases.\\

This can be achieved through the alignment algebra introduced in section \ref{key},
then we formally define a composed alignment, denoted by $\mathcal{A}_{ik}$, as follows :

\begin{definition}

Given two alignments $\mathcal{A}_{ij}$ and $\mathcal{A}_{jk}$, if it exists a bridging
entity interlinking two existing direct correspondences, then the
composed alignment $\mathcal{A}_{ik}$ is defined by a set of composed correspondences :

\[\mathcal{A}_{ik}= \mathcal{A}_{ij} \oplus \mathcal{A}_{jk} \]
\vspace{-0.5cm}
\[\mathcal{A}_{ik}= \{\langle e_1, e_3, \mathcal{F}_{rel}(r,r'), \mathcal{F}_{conf}(n,n')\rangle | e_1 \in \mathcal{O}_{1}, e_3 \in \mathcal{O}_{3} \} \]

where :

\begin{itemize}

  \item $e_1 \in \mathcal{O}_{1}$, $e_2 \in \mathcal{O}_{2}$ and $e_3 \in \mathcal{O}_{3}$.
  \item $\langle e_1, e_2, r, Conf(n) \rangle \in \mathcal{A}_{ij}, \langle e_2, e_3, r', Conf(n')\rangle \in \mathcal{A}_{ij}$.
  \item the bridge entity $e_2$.
  \item $\mathcal{F}_{rel}$, $\mathcal{F}_{conf}$, are functions designed for composing two relations and two
  confidence values, respectively.
\end{itemize}
\end{definition}

In the following, a detailed description of the sub
modules constituting the indirect
alignment composition module.

\subsubsection{Bridge Entity Detection} \label{bridge}

\medskip

In this step, from developed parsed entity lists obtained
by the \emph{parsing and paths identification} module,
we define the most important element of the composition,
which is the \emph{bridge entity}. This \emph{bridge entity} ensures and guarantees navigability
as well as semantic transitivity between any three alignments (\emph{i.e.}, previously identified
by the \emph{storage examiner} component).
The \emph{bridge entity} detection can involve several techniques,
from string based similarities to the semantic test treatments, including any possible external
resource (these aspects are discussed in detailed manner through the experimental study).
In our context, each selected alignment will be parsed and transformed into a list of
entity couples. This sub-module is carried out according to techniques
manifested through terminological and syntactic similarity.

\subsubsection{Relation Composition} \label{RelComp}

New relation composition that label the link between two given entities $e_1$
and $e_3$, as previously described is achieved through the composition table and
presented in the sub-section \ref{key}. For example, in the case of the equivalence relation denoted as $\equiv$, if $e_1 \equiv e_2$, $e_2 \equiv e_3$ then $e_2 \equiv e_3$ and the new relation will characterize the fact that $e_1$ is equivalent to $e_3$.

\subsubsection{Confidence Value Computation} \label{RelComp}

Confidence value composition is performed according to one of
four manners (\textit{i.e.} namely, multiplication, normalization, maximization and minimization) introduced in section \ref{key}.

\subsection{Refinement and enrichment}

The purpose of the refinement and enrichment module is to improve the
result of the composition through the extension and semantic transitivity
of an entity that is part of a product alignment. In a first step, the
detection of the bridge entity was based on a simple equality test guided
by a syntactic similarity measure. In a second step, and through this
module, we seek to increase the chance of detection of the bridge entity
through the use of an online dictionary.

\subsection{Indirect Alignment Generation} \label{Iag}

The alignment file generation is the last step in the \textsc{Cimona}
method process. At this level, an alignment file is built from the aligned
entities, as well as the relation and its confidence value.
This step is performed by means of the alignment API\footnote{http://alignapi.gforge.inria.fr/}.

\section{Experimental Study} \label{etude}

In what follows we will present the experimental study,
based on the metrics of Precision\footnote{$Precision=\frac{\mid N_{correct} \mid}{\mid N_{found}\mid}$} (P), Recall\footnote{$Recall=\frac{\mid N_{correct} \mid}{\mid N_{expected} \mid}$} (R).
Subsequently, a discussion is presented according
to the internal characteristics of the considered methods output.

\subsection{Test Cases}

The carried out experimental evaluation uses alignments file-battery provided by the OAEI campaign (Ontology Alignment Evaluation Initiative Campaign). These files are alignment methods outputs that participated in the 2020 campaign. Tests were conducted on alignments produced by three pioneering methods of OAEI campaign in Multifarm\footnote{http://oaei.ontologymatching.org/2020/results/multifarm/index.html} track, namely Aml \cite{AML2020}, LogMap \cite{LogMap2020} and LogMapLt \cite{LogMap2020}. Indeed, the choice of this test base is well studied, considering its evaluation specificity. This track consists of finding alignments between translated ontologies of a subset of the Conference track, translated in nine different languages (\emph{i.e.}, Arabic (ar), Chinese (cn), Czech (cz), Dutch (de), French (fr), German (nl), Portuguese (pt), Russian (ru), and Spanish (es)).

\subsection{Results and Discussion}

The results are presented according to two levels, aiming
at highlighting the characteristics of the proposed method.
The first scenario examines alignment aspects in intra-method,
\emph{i.e.}, according to the particularities of each alignment
method. While the second scenario deals with the composition of the alignments files resulting as outputs of three different methods.
Indeed, such a choice of experimentation
is guided by the need to examine the sensitivity of the composition
process as to the operating details of the direct alignment
methods. Moreover, for each studied scenario, we have composed
the confidence value according to the $4$ possibilities
cited in subsection \ref{compcomp}.\\

Table \ref{tab1} summarizes the Precision and  Recall values obtained for each alignment method (from row 1 to row 5 we computed the confidence value using maximization, from row 6 to row 10 using minimization, from row 11 to row 15 using multiplication and finally from line 16 to line 20 using normalization). Indeed, from a couple of considered alignments we were able to deduce a new composed one. The \textsc{Aml} method takes a net ascendancy compared to other methods. This argues that the quality
of the indirect alignment is closely related to the quality
of the direct alignment. In other words, the efficiency
of the \textsc{Aml} method in direct alignment explains its good
performance during the composition process, and the good
quality of its direct matching has clearly and conspicuously
marked the composition process, leading to good values of Precision and Recall against reference alignment.\\

\begin{table}
\centering
\caption{Evaluation metrics according to confidence value computation scenarios}\label{tab1}
\begin{tabular}{{|p{0.5cm}|p{2.5cm}|p{3cm}|p{1.5cm}|p{1.5cm}|p{2cm}|}}
\hline
     &   \textbf{Direct Couple}  &  \textbf{Indirect Couple}     & \textbf{\textsc{Aml}} & \textbf{\textsc{LogMap}}   & \textbf{\textsc{LogMapLt}} \\  \hline \hline

\textbf{1}  &  fr-en/en-es   &  fr-es     &   0.74/0.56   & 0.68/0.40  &  0.00/0.00    \\ \hline
\textbf{2}  &  it-fr/fr-pt   &  it-pt     &   0.74/0.60   & 0.60/0.30  &  0.90/0.15    \\ \hline
\textbf{3}  &  it-fr/fr-nl   &  it-nl     &   0.70/0.50   & 0.74/0.35  &  0.80/0.04   \\ \hline
\textbf{4}  &  de-pt/pt-ru   &  de-ru     &   0.66/0.40   & 0.76/0.40  &  0.00/0.00    \\ \hline
\textbf{5}  &  en-it/it-pt   &  en-pt     &   0.76/0.48   & 0.74/0.50  &  0.82/0.05     \\

\hline
\hline

\textbf{6}  &  fr-en/en-es   &  fr-es     &   0.72/0.52   &  0.64/0.36  &   0.00/0.00   \\ \hline
\textbf{7}  &  it-fr/fr-pt   &  it-pt     &   0.70/0.58   &  0.58/0.28  &   0.85/0.10   \\ \hline
\textbf{8}  &  it-fr/fr-nl   &  it-nl     &   0.70/0.49   &  0.70/0.30  &   0.78/0.03   \\ \hline
\textbf{9}  &  de-pt/pt-ru   &  de-ru     &   0.63/0.37   &  0.70/0.36  &   0.00/0.00   \\ \hline
\textbf{10} &  en-it/it-pt   &  en-pt     &   0.72/0.42   &  0.70/0.48  &   0.78/0.03   \\

\hline
\hline

\textbf{11}  &  fr-en/en-es   &  fr-es     &  0.42/0.22    & 0.32/0.14   &  0.00/0.00    \\ \hline
\textbf{12}  &  it-fr/fr-pt   &  it-pt     &  0.38/0.20    & 0.20/0.08   &  0.55/0.05     \\ \hline
\textbf{13}  &  it-fr/fr-nl   &  it-nl     &  0.38/0.18    & 0.42/0.22   &  0.44/0.01    \\ \hline
\textbf{14}  &  de-pt/pt-ru   &  de-ru     &  0.43/0.18    & 0.42/0.24   &  0.00/0.00    \\ \hline
\textbf{15}  &  en-it/it-pt   &  en-pt     &  0.42/0.16    & 0.42/0.28   &  0.43/0.01    \\

\hline
\hline

\textbf{16}  &  fr-en/en-es   &  fr-es     &  0.22/0.04    &  0.18/0.04  &  0.00/0.00   \\ \hline
\textbf{17}  &  it-fr/fr-pt   &  it-pt     &  0.18/0.02    &  0.06/0.00  &  0.00/0.00   \\ \hline
\textbf{18}  &  it-fr/fr-nl   &  it-nl     &  0.18/0.03    &  0.20/0.02  &  0.00/0.00   \\ \hline \textbf{19}  &  de-pt/pt-ru   &  de-ru     &  0.23/0.05    &  0.20/0.04  &  0.00/0.00   \\ \hline
\textbf{20}  &  en-it/it-pt   &  en-pt     &  0.22/0.06    &  0.20/0.02  &  0.00/0.00   \\ \hline

\end{tabular}
\end{table}

The \textsc{Aml} method is succeeded respectively by \textsc{LogMap} and \textsc{LogMapLt}. The values for the first two scenarios; are very close. This is due to the fact that the confidence
values deduced by the alignment methods are close and do not mark
any degradation which influence the rate of correctly aligned
feature pairs. On the other hand, the values decrease considerably
in the two cases of composition of the confidence value, \emph{i.e.},
multiplication and normalization. This is due to the fact that
these two methods of composition attenuate the confidence
values which make them outside the range of the reference alignments.\\

As a general observation, we conclude that the direct alignment method that leads to good results is an asset to initiate the alignment
composition phase in order to deduce composed alignments
and leads to good results. This observation is confirmed
by the \textsc{Aml} method, even for the two confidence
composition mismatches that attenuate its performance.
Moreover, maximization and minimization approaches are best suited
for composition by preserving a good level of semantic
similarity propagated through the composition process,
leading to good results. The same remarks are confirmed
by the remaining outputs of the others methods,
even though they have less performance.\\

It should be mentioned that the performance of an indirect alignment
method is closely related to the quality of the direct alignment
used as well as the method of composition of the confidence value.
This leads us to exploit the outputs of the pioneering methods in
direct alignment by putting in place good composition mechanisms
that value and capitalize the direct stored alignments on the one
hand and that respond to the needs of applications and programs
that exploit them.

\section{Conclusions} \label {conclusion}

This paper introduced the \textsc{Cimona} method for indirect cross-lingual
ontology alignment. The obtained results are very promising
and accentuate other aspects related to ontology alignment and reuse.
Through a thorough experimental study, we investigated the determining
and influential factors of the alignment composition task. Our motivation
is to show the feasibility and the realization of such an approach with
the aim of interconnecting some existing isolated data sets, especially
in a vital domains. Indeed, the creation and deduction
of new alignments by composition, creates semantic bridges to promote
reasoning and facilitate the tasks of excavation, interrogation, mediation etc.\\

The proposed method showed a good performance,
but still requires some improvements. In the near future, we also intend
to enhance the performance of the \textsc{Cimona} so that it can
handle a wider range of alignments and under other logical constraints and expressions. Furthermore, the integration of new external
resources can provide a wider choice of semantic equivalents that could be of benefit to the task of \emph{bridge entity} detection.
Besides, a Graphical User Interface (GUI) is needed to assist ordinary users and make the composition process interactive. In addition, we work on the refinement of
semantic relation composition, \emph{e.g.},
logical links are usually dependent of ontology development formalism details and languages.
Doing so, we have to investigate the impact of other logical and
semantic operators composition, especially in the case of highly dense hierarchy with
high granularity degree of specification and/or certain domain description.
In addition, we plan to integrate the \textsc{Cimona} method to
a complete ontology mediation system also in interactive environments.

\end{document}